\documentclass{article}

\usepackage{microtype}
\usepackage{graphicx}
\usepackage{subfigure}
\usepackage{booktabs} 
\usepackage{enumitem}
\usepackage{amsmath}
\usepackage{amssymb}

\usepackage{hyperref}
\usepackage{placeins}

\usepackage[accepted]{icml2020}

\icmltitlerunning{}

\begin{document}

\newcommand{\eye}{\boldsymbol{I}}
\newcommand{\Apvar}{A^0_{1:n_p}}
\newcommand{\tpvar}{t^0_{1:n_p}}
\newcommand{\Acvar}{A^{k+1}_{1:n_c}}
\newcommand{\tcvar}{t^{k}_{1:n_c}}
\newcommand{\scvar}{s^{k}_{1:n_c}}
\newcommand{\Aivar}[1]{A^{#1}_{\cdot}}
\newcommand{\tivar}[1]{t^{#1}_{\cdot}}
\newcommand{\sivar}[1]{s^{#1}_{\cdot}}
\newcommand{\latentvar}{\tpvar, \Apvar, \scvar}
\newcommand{\obsvar}{\tcvar, \Acvar}
\newcommand{\expect}[2]{\mathbb{E}_{#1}\left[#2\right] }
\newcommand{\kl}[2]{D_{KL}\left({#1}:{#2}\right)}
\newcommand{\lewis}[1]{\textcolor{red}{#1}}
\newcommand{\lisa}[1]{\textcolor{blue}{#1}}
\newcommand{\tk}[1]{\textbf{TK:} {\color{red} #1}}
\twocolumn[
	\icmltitle{Capsule Networks---A Probabilistic Perspective}

	\icmlsetsymbol{equal}{*}

	\begin{icmlauthorlist}
		\icmlauthor{Lewis Smith}{ox}
		\icmlauthor{Lisa Schut}{ox}
		\icmlauthor{Yarin Gal}{ox}
		\icmlauthor{Mark van der Wilk}{imp}
	\end{icmlauthorlist}
	\icmlaffiliation{ox}{OATML Group, Department of Computer Science, University of Oxford, United Kingdom}
	\icmlaffiliation{imp}{Department of Computing, Imperial College London, United Kingdom}

	\icmlcorrespondingauthor{Lewis Smith}{lsgs@robots.ox.ac.uk}

	\icmlkeywords{Machine Learning, Probabilistic Modelling, Graphical Models, ICML}

	\vskip 0.3in
]

\printAffiliationsAndNotice{}  

\begin{abstract}

`Capsule' models try to explicitly represent the poses of objects, enforcing a linear relationship between an object's pose and that of its constituent parts. This modelling assumption should lead to robustness to viewpoint changes since the sub-object/super-object relationships are invariant to the poses of the object. We describe a probabilistic generative model which encodes such capsule assumptions, clearly separating the generative parts of the model from the inference mechanisms. With a variational bound we explore the properties of the generative model independently of the approximate inference scheme, and gain insights into failures of the capsule assumptions and inference amortisation. We experimentally demonstrate the applicability of our unified objective, and demonstrate the use of test time optimisation to solve problems inherent to amortised inference in our model.

\end{abstract}

\section{Introduction}
\label{sec:intro}
Capsule models \citep{sabour2017dynamic,hinton2018matrix} attempt to use the knowledge that large-scale objects are composed of a geometric arrangement of simpler objects. In rigid bodies, the poses of child-objects are related to the whole object through an affine transformation. Relative poses are represented using a matrix $M$ in homogeneous coordinates, which allows the pose of child-objects to be expressed as $A_c = A_p M$. Using this linear relationship in pose allows models to 1) be more insensitive to changes in poses, due to the strong generalisation of linear relationships, and 2) use consistency in pose of features as an additional check when predicting the presence of an object.

This intuition behind capsules stems from a generative argument: if we know that an object is present, then the child-objects it is composed of must be present and have consistent poses that are predicted through the linear relationship above. When inferring the presence of an object, this reasoning is inverted. The pose of each child object predicts a pose for the composite object through $A_p = A_c M^{-1}$. An object is likely to be present if the child-objects it is composed of are present \emph{and predict similar poses} for the main object. If the poses are not consistent, then child-objects are unlikely to have been caused by the presence of a single object.

\citet{sabour2017dynamic} and \citet{hinton2018matrix} proposed \emph{routing} algorithms for finding consistent poses. The latter finds clusters of consistent poses from child-objects using a procedure inspired by Expectation Maximisation (EM) for Gaussian mixtures. 
These routing methods are explained as performing inference on some \emph{latent variables} in the model, but are not derived from a direct approximation to Bayes' rule.

In this work, we express capsule networks within the probabilistic framework, showing how a `routing' algorithm can be derived in this context. We do so with two main aims. Firstly, we aim to express the modelling assumptions of capsules as a probabilistic model with a joint distribution over all latent and observed random variables.
This explicit expression through probability distributions and conditional independence allows the model specification to be critiqued and adjusted more easily.
Our second aim is to provide a principle by which algorithms for taking advantage of these assumptions can be derived. By phrasing the problem as approximate inference in a graphical model, we have a procedure for deriving routing schemes for modified models, or for  taking advantage of the literature on inference in graphical models.

Our contributions are as follows:
\begin{enumerate}
  \item We express the `capsule' assumptions as a generative probabilistic model.
  \item We derive a routing method from clear variational inference principles, leading to an amortised method similar to variational autoencoders \citep{kingma2013auto,rezende2014dlgm}.
  \item We show comparable empirical results to previous work on capsules, showing that our probabilistic approach does not inadvertently lose existing capsule properties.
\item We investigate whether the inference method reflects the assumptions in the generative process, namely the
degree to which capsules specialise to objects and capture object pose in an equivariant way if exposed to variation in pose at \textit{training time}, demonstrating limitations of current approaches. We discuss the implications of our probabilistic framing and our empirical findings for future work.

\end{enumerate}

\section{Capsule Networks}
\subsection{A graphical model}
\label{sec:caps}
Previous work \citep[e.g.~][]{kosiorek2019stacked,hinton2018matrix} describe isolated parts of their capsule models using likelihoods. While these likelihoods inspire loss terms in the training objective, the likelihoods do not form a complete generative procedure over all variables, and the distinction between latent (random) variables and neural architecture choices is not made. From the recent approaches to capsules, \citet{kosiorek2019stacked} rely most on expressing loss terms as likelihoods but lack a generative expression, for example, over the presences of child objects. 
We intentionally follow the model structure and intuitions of earlier work but take care to build a probabilistic model with a globally well-defined joint probability distribution over all random variables, allowing us to derive many possible inference methods for the same graphical model.

It is important to note that previous iterations of capsules have used
different models, some fully discriminative, and some
with a partial or fully unsupervised autoencoding framework.
We discuss connections to prior work in more detail in Section \ref{sec:related}. 

The assumptions that capsules try to capture about a scene are
\begin{enumerate}  
    \item The scene can be described visually in terms of a set of simple objects, or graphics primitives
    \item The positions of the simple objects can be described concisely in terms of complex objects (parents), which are geometrical arrangements of simpler objects (children)
    \item Objects can be organised into a hierarchy, with simple objects at the bottom and complex objects higher in the hierarchy
    \item Each child has at most one immediate parent---that is, it is a part of at most one higher-level object, although we may be uncertain about which one this is.
\end{enumerate}

We introduce a generative model that captures these intuitions. Different images have their own composition of objects, and therefore have image-specific random variables representing this. In the following, per-image random variables are in black, while global parameters that are shared across different images are denoted in red. Each object that can be present in an image is represented through its presence $t$ and pose $A$. Objects are present when $t=1$, and absent otherwise. A capsule is the presence and pose for an object taken together. To form the hierarchy of objects, we arrange capsules in layers denoting the layer $k$ in the superscript, and the object $i$ in the subscript, i.e.~$t^k_i$. 
The top layer capsules are independently present at a random pose, encoded as
\begin{align}
    t^0_{i} &\sim \textrm{Bern}(p)\,, & 1 \leq i \leq n_0\,, \\
    A^0_{i} &\sim \textrm{Norm}(0, I)\,, & 1 \leq i \leq n_0\,.
\end{align}
Each parent object is likely to be composed of a different combination of child objects, so we represent the affinity of a parent $i$ to a child $j$ using the global parameter $\color{red}\rho^k_{ij}$. In any instantiation, a child is part of a single object, which is represented with the selection variable $s$:
\begin{align}
    s^k_{j}\!\mid\!\left\{t^{k-1}_i\right\}_{i=1}^{n_p} \!\sim \textrm{Cat}\!\left(\frac{[..., {\color{red}\rho^k_{ij}} t^{k-1}_i,...]}{\sum_{i}^n {\color{red}\rho^k_{ij}} t^{k-1}_i}\right) , 1 \!\leq\! j \!\leq\! n_k .
\end{align}
A capsule that is not on (i.e.~$t^{k=1}_i = 0$) has zero probability of being selected by a child. Note that we include a dummy parent capsule that does not correspond to an object, does not have a parent of itself, and is always on. This allows capsules to not be a part of an object, so ``clutter'' can be accounted for. This is also necessary in order to have a well defined probability if all parents are off.

The actual presence of the child, conditioned on its parent, occurs with probability $\color{red}\gamma^k_{ij}$:
\begin{align}
    t_{j}^k &\mid \left\{s^k_j = i\right\} \sim \mathrm{Bern}\!\left({\color{red}\gamma_{ij}^k}\right) \,, & 1 \!\leq\! j \!\leq\! n_k\,.
\end{align}
If a child is on, its pose is given by the selected parent pose multiplied by a learned pose offset $\color{red}M^k_{ij}$. By using a Gaussian likelihood with a learned variance $\color{red}c_{ij}^k$, we allow for some mismatch between poses in individual images. If a child is switched off, it plays no part in image generation. However, we still represent the pose, just with a large variance $\color{red}\lambda_{\text{off}}$.\footnote{It is necessary to do this because we will need to relax the discreteness of the parents to make inference tractable.} This gives the conditional distributions:
\begin{align}
    A^k_j \!\!\mid\!\! s^{k}_j\!=\!i, t^{k}_{i}\!=\!1, A^{k-1}_{i} & \!\sim \textrm{Norm}((I \!+\! A^{k-1}_j) {\color{red} M^k_{ij}}, {\color{red} c^k_{ij}})  \\
    A^k_j \!\!\mid\!\! s^{k}_j\!=\!i, t^k_{i}\!=\!0, A^{k-1}_{i} & \!\sim \textrm{Norm}(0, {\color{red}\lambda_{\textrm{off}}}I)) \,.
\end{align}
Notice we add the identity to $A$---this is primarily for convenience, as it lets us parameterise $A$ as normally distribution variables centred around zero, rather than the identity.

When accounting for clutter by attaching to the dummy capsule, the child is given a random pose, which is not generated to be consistent with other capsules. The dummy capsule is given the zero index in each layer, and child capsules have a fixed small probability of attaching to it (implemented by ${\color{red}\rho^k_{0j}} > 0$). A random pose is generated by setting a large $\color{red}c^k_{0j}$.

\begin{figure}
    \centering
    \includegraphics[width=0.8\columnwidth]{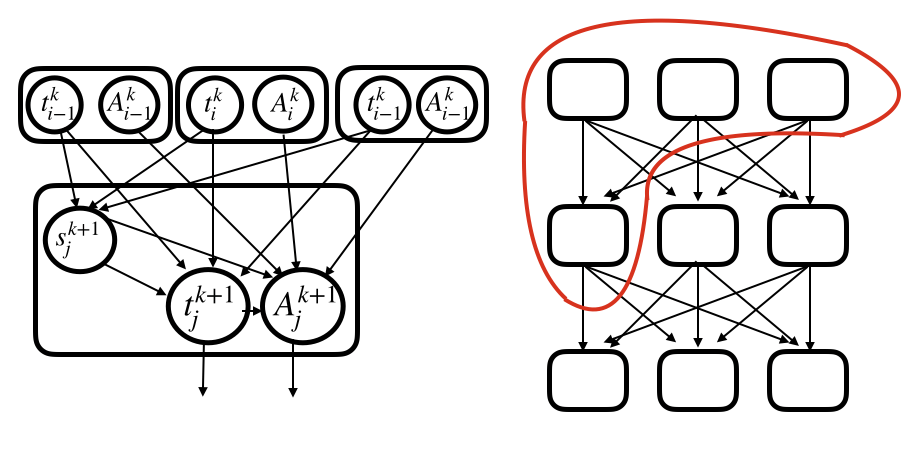}
    \caption{A sketch of the generative model for a capsule network. The broad connectivity pattern is shown on the right, and the detailed connectivity between the random variables is shown on the left, for the region of the overall graph circled in red. }
    \label{fig:capsules}
\end{figure}

Figure \ref{fig:capsules} shows a graphical representation of the conditional distributions between a capsule and the parent layer, as well as how capsules are arranged in a network.

So far, we have defined our model for relating the poses of complex objects to simpler ones.  However, our final output is the pixels in images, not a collection of objects and poses.  As a result, we need to relate the capsules in the final layer $K$ (representing the
simplest objects) directly to the pixels of the image. A simple way of doing
this, following \citet{tieleman2014optimizing,kosiorek2019stacked}, is to associate a template $T_j$ and alpha channel template $\alpha_j$ with every lowest level capsule.  To build up an image,  we transform the templates by their respective poses $A^K_j$, and then build up the image encoded by these using alpha compositing, drawing them in an arbitrary sequence, with the alpha channels weighted by the presence probability of that object.
\begin{align*}
  \textrm{\textbf{alpha\_compositing}}&(T_{1:n}, A^K_{1:n}, t^K_{1:n}, \alpha_{1:n}):\\
  P &\gets \textrm{zeros} \\
  &\text{for}\; 1..n_{K} \\
  P &\gets \text{over}(A^K_i T_k, t^K_i * A^K_i \alpha_i, P) \\
  \text{return}(P)
\end{align*}
If we have two images $C_a$ and $C_b$, with alpha channels $\alpha_a$ and $\alpha_b$, then the over operation is 
\begin{align*}
    C_o &= \frac{C_a \alpha_a + C_b \alpha_b ( 1 - \alpha_a)}{\alpha_o}\\
    \alpha_o &= \alpha_a + \alpha_b ( 1 - \alpha_a)
\end{align*}
performed in parallel over each channel and pixel.
This, in effect, draws $C_a$ on top of $C_b$, accounting for the transparency of $C_a$ encoded by the alpha channel.
In order to have a data likelihood with full support in image space, we need to add noise to the data distribution.
We use Gaussian noise, so the likelihood for an image $X$ conditioned on the lowest level capsules is then
\begin{align}
  X_{ij} \mid t^K_{1:n}, A^K_{1:n} \sim \textrm{Norm}(P_{ij}, \sigma) \,,
\end{align}
i.e.~an independent Gaussian on each pixel centred on the mean of the composited image.
Drawing the templates in this way has the side-effect of introducing an
arbitrary ordering of the templates.  This is somewhat undesirable, since it has no
`physical' basis, as the ordering of the templates is an internal detail of the
model, but the template model is somewhat artificial in any case.

While our use of templates is shared by the recent approach by \citet{kosiorek2019stacked}, we do believe this is a weak point in the current formulation of the model. Template rendering is much more tightly prescribed and inflexible rendering method than a neural network, like what is used in the variational autoencoder \citep[VAE ---][]{kingma2013auto,rezende2014dlgm}. However, using such a flexible rendering model would introduce non-identifiability between poses explained by the capsules, and by the neural network, as the variation in the appearance of the object due to variations in pose can also be explained by the neural component of the decoder.
In order to focus on evaluating the probabilistic model and inference in it, we chose to follow the more interpretable existing template methods \citep{tieleman2014optimizing}.

\subsection{Inference}

Image processing in our model consists of inferring the posterior over the latent variables conditioned on the image. Further, we need to perform inference in order to fit the generative model, as the global parameters
of the generative process are not known in advance. The standard approach in
deep generative modelling is to use variational inference with the reparameterisation trick, but this is complicated in this generative model by the presence of discrete random variables. However, due to the dependency structure of
the model, the selection variables $s$ are conditionally independent given all other variables in the model (see Figure \ref{fig:capsules}).
That is, given $A^k, t^k, A^{k-1}, t^{k-1}$, the $s_i^k$ are all conditionally independent.
As such, these can be marginalised out analytically for a cost in space and time of the order of the number of parent capsules in a layer.\footnote{This also avoids applying Jensen's inequality for these variables in the derivation of the variational bound.}

\begin{figure}[H]
    \centering
    \includegraphics[width=\columnwidth]{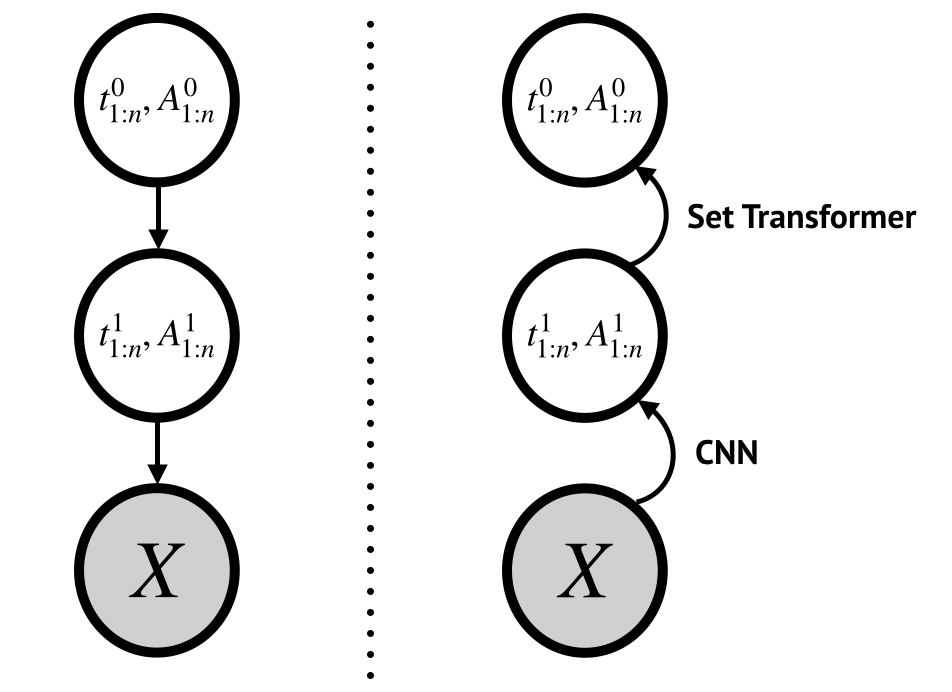}
    \caption{Left: the high level dependency structure in the generative model. Right: the dependency structure in the approximate posterior; variables are sampled from the approximate posterior at each layer before being passed on to the next layer, so the dependency structure of the amortised posterior is inverted compared to the generative setup ($t^0, A^0$ depend on $t^1, A^1$, rather than the other way around).
    }
    \label{fig:inference_diagram}
\end{figure}

This cannot be done for the presence variables $t$, since the distribution of
each $t$ depends on \textit{all} its potential parents $t_1, t_2, ... t_n$,
and so enumeration would have a cost exponential in the number of parents in
the model.  Even for the relatively small numbers of capsules considered in this paper, this would be impractical, so
instead we modify the model to relax the $t$ variables, replacing them with
concrete distributed variables \cite{maddison2016concrete,jang2016categorical}.
With these changes, we can derive a tractable Monte Carlo estimator of the
evidence lower bound for an approximating distribution over $A$ and
the relaxed $t$ variables. 
In our inference network, we reverse the dependency structure of the generative model, as shown in Figure \ref{fig:inference_diagram}, parameterising $p(t^1, A^1)$ with a CNN, and $p(t^0, A^0)$ with a set transformer (more details on this are given in section \ref{sec:arch}.)
After enumeration of the $s$, the lower bound is then of the form

\begin{align}
  \mathcal{L}(\theta, \phi) &= \mathbb{E}_{q(t, A; \phi)}\left[\log p(t, A; \theta) - \log q(t, A; \phi)\right]\\
  &=\mathbb{E}_{q(t, A; \phi)}\Big[ \log (\sum_{s}[p(s \mid t, A) p(t, A, s)]) \nonumber\\&\hspace{4em} -\log q(t, A)\Big] \nonumber
\end{align}

In our code we do not need to derive the full bound manually, as we use a bound derived automatically by expressing our model in the probabilistic programming language Pyro \cite{bingham2018pyro}.
As a result, we omit it from the main body of the paper, but we include a full derivation in the supplementary material. 

We wish to perform amortised inference \cite{kingma2013auto} for
efficiency.  This is partly for fast inference at test time, but mostly
because, as in a VAE, we also want to fit the global parameters of the
generative model by maximising the marginal likelihood. Using an expensive
inference procedure would likely be too slow to make this practical.

\subsection{Approximating distributions}

In principle, we should be able to use any approximating distribution in our variational bound.
In practice, training the model from initialisation with a full variational bound is extremely difficult. 
Since we cannot take expectations explicitly in the bound, we approximate it using sampling.
The variance from this tends to mean that the model underfits if we train from this from the start, as near the start of training there is no real signal encouraging the variational posterior to be different from the prior, but if the variational parameters are sampled from the prior, there is also no signal leading to specialisation of the templates.
In addition, the likelihood over the template pose is extremely strongly peaked, with a large likelihood for a `correct' placement, but with large penalties for small deviations from this, which makes the problems arising from the variance much more severe.

In order to get around this problem, we use deterministic poses at least during the initial training of the model, leading to a partially variational bound, where the discrete variables are marginalised out, the presence variables are approximated with concrete distributions, and the poses are approximated with Delta distributions, so we perform MAP inference over poses rather than finding a full variational approximation to their posterior, at least during training.

It is important to note that we verify that this is not a theoretical issue with the bound so much as it is a practical issue with training.
Starting from a model initialised with the EM parameters, we are able to train an encoder with a full variational posterior, achieving comparable results to training with deltas. 
This achieves a tighter ELBO than training a model with the variational bound (with random sampling) from initialisation.
Unless otherwise noted, however, we report results with delta poses in section \ref{sec:related}, as this is faster to train and we wanted to check a large variety of configurations, and we did not find that this re-training with the full ELBO had much effect on the metrics we use.

\subsection{`Free form' variational inference}

\begin{figure}[h]
  \centering
  \includegraphics[width=\columnwidth]{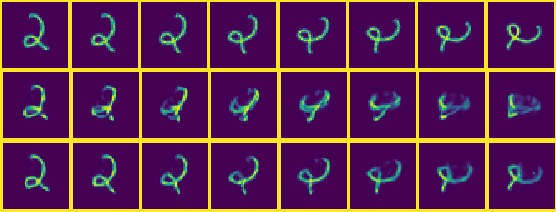}
  \vspace{-0.3cm}
  \caption{
  Top row: MNIST digit, with rotations applied.
  Middle row: Amortised reconstructions by a capsule model. Note the failure for some rotations.
  Bottom row: Reconstructions after free (test time) optimisation of our variational bound, showing that our unified objective is meaningful.
}
\label{fig:rotation_visualisation}
\end{figure}

As well as amortised inference, it is possible to perform standard variational inference at test time using our objective.
That is to say, rather than having the variational parameters $\phi$ be the outputs of a neural network, we instantiate a set of $\phi$ for each image, and optimise them separately using gradient descent.
The analytic marginalisation of the $s$ variables in our bound makes this correspond to an EM algorithm - we sample the continuous random variables from our variational distribution $q(\phi)$, then marginalise the discrete variables $s$ conditioned on these analytically, and iterate these alternating steps.

This allows us to investigate the properties of our variational bound independently of our amortisation network.
The advantages of this are demonstrated in Figure \ref{fig:rotation_visualisation}.
This shows the results of reconstructing an out of distribution image, a rotated 2, using a generative model and inference network trained on standard MNIST. The encoder is able to generalise to small unseen rotations, but begins to fail for large rotations, which are very unlike the data it saw during training.
However, if at test time we continue to optimise the variational parameters freely, initialising them at the values predicted by our encoder network, then the quality of reconstruction markedly improves, even for out of distribution images, as the \textit{generative} model assigns a reasonable likelihood to these out of distribution images. We find also that optimising the bound at test time, initialising at the output of the inference network, before classifying based on the latent variables (see below) can improve classification accuracy based only on $t^p$ by a modest amount. If we take a trained model and optimise the latent variables with respect to this bound, starting from the amortised model output, then we can improve the classification results based on $t^0$ quoted in section \ref{results} by a small amount, about $0.5\%$. The fact that we can do optimisation at test time and it improves the performance shows that our probabilistic bound is a useful and consistent objective. However, we do not report this in our results below, as it is extremely inefficient\footnote{This is because the gradient updates are slow to converge, so we need to take hundreds or thousands of steps to see real improvement in the bound, which means a significant amount of time spent per example, as each gradient step requires a pass through the generative model to calculate the gradient with respect to the variational parameters. If we take 1000 steps of gradient descent, which is not too unreasonable, then it takes 1000 times longer to process a single digit, which is hardly a fair comparison to other methods and would have made testing any sensible number of configurations prohibitive. It may be possible to improve this speed a lot by using a second order optimiser but we did not explore this here.} to optimise this bound at test time using basic gradient descent relative to the forward pass of a neural network.

\section{Related Work}
\label{sec:related}

There have been many different models which have been called
`capsules' over the years, with the idea dating back to at least
\citet{hinton2000learning}.  Interestingly, \citet{hinton2000learning}
used an explicit graphical modelling perspective, specifying a
generative process over images, though not one that explicitly
included poses.
Similar hierarchical, graphical approaches were explored by \citet{storkey2003image}, though their model did not incoporate the pose offset (our $M^k_{ij}$) used in capsules.
However, much subsequent work explored the ideas
of having a vector activation in a purely discriminative model (with various
training schemes), including explicitly geometrical
\citep{hinton2011transforming}, and purely discriminative, but with
an unusual routing mechanism \citep{sabour2017dynamic, hinton2018matrix}.

Other routing approaches have been proposed, but we focus here on a probabilistic treatment of capsules, in particular based on \citet{hinton2018matrix}, so we do not discuss these further. We focus on the view of capsules as representing objects specifically, so we do not consider the broader interpretation of capsules as meaning any model with vector valued activations which is sometimes used in the literature.

It is worth noting, however, that the routing algorithm of
\citet{hinton2018matrix} is explicitly derived as an approximation to an EM
procedure for a particular graphical model.  The probabilistic model of a `mixture of switchable transforming Gaussians' is not explicitly written down in \citet{hinton2018matrix} (its variational free energy is given). However, the graphical model between a set of parent poses and activations,
and child poses and activations given in Section \ref{sec:caps} corresponds closely to it. In \cite{hinton2018matrix}, however, this inference problem is used as a black
box activation function in a model that is trained in a discriminative
way. In contrast to the earlier routing algorithm of
\cite{sabour2017dynamic}, both steps of the routing algorithm of
\cite{hinton2018matrix} minimise the same objective function.  However, this
objective function is local to a layer, and as the overall algorithm is trained
to minimise a discriminative objective, it has no particular relationship to the overall loss. 
In contrast, making the model generative means that the \textit{entire model} minimises a \textit{single}, consistent variational objective.

Approaches to capsules that have been trained discriminatively have often included a reconstruction term in the loss, for example, \citet{sabour2017dynamic} and \citet{rawlinson2018sparse}.
In these works, however, the generative component of the model was a neural decoder that took the top-level capsule vectors as input but treated them as an unstructured latent space.
In contrast, our generative model reflects the assumption that the latent variables in the model explicitly represent poses.

The decoder of the autoencoder model in the recent work of
\citet{kosiorek2019stacked}, is conceptually close to the generative model presented in this paper, and we use architectures based on the encoder described in their paper to perform inference. However, we adopt a more parsimonious approach with fewer templates and variables, contrary to \citet{kosiorek2019stacked} who use a more complex decoder--- with more capsules, additional sparsity losses, and features like deformable templates and special features. Although these may allow for better performance, our main focus in this work is introducing a probabilistic treatment and analysing the model in this framework, rather than achieving state of the art performance.  It is important to note that we find that the model described in \citet{kosiorek2019stacked} and our model suffers from a similar pathology when training on augmented data, which we describe in more detail in the next section.

Further, as mentioned before, the variables in this model are not explicitly modelled as random, and there is no consistent joint probability defined over all variables in the model, despite descriptively referring to activations in the model as probabilities. The different components of the model being trained to maximise an image reconstruction likelihood and the higher-level part of the model maximising the log-likelihood of part poses, with a stop gradient between these two components (apart from on `special features', which play an unclear role in generation). 
In contrast, we introduce a fully probabilistic view, which allows a clear separation of the \textit{generative} modelling of capsule assumptions from the inference technique.

\section{Model Architecture}
\label{sec:arch}

In the following results, we use a relatively small model configuration of two layers, consisting of $16$ higher-level capsules and $16$ low level (template) capsules. We experimented with varying this but were unable to find a hyperparameter setting, which produced much better results, though we did not perform extensive tuning.
We use a template size of $12$. This means our generative model has very few learnable parameters compared to a neural model like a VAE decoder, with fewer than $30,000$ parameters. For comparison, even a simple VAE mapping from a $10$-dimensional latent space to MNIST with $128$ hidden units has over $100,000$ parameters.

For the CNN which predicts $t^1, A^1$ conditioned on the image, we use a $4$ layer CNN, with a stride of $2$ in the first two layers, and $32$ channels. We use the attentive pooling mechanism described in \citet{kosiorek2019stacked}.
For our set transformer, which predicts $t^0, A^0$, we use $32$ inducing points and $4$ attention heads, with a hidden dimensionality of $128$, across $4$ ISAB layers followed by PMA (see \citet{lee2019set}). After the PMA layer, we use small MLPs to map the resultant $32$ dimensional hidden state to the logits of the $t^0$ and the mean of the $A^0$ distribution (when using delta distributions in the variational/EM objective).
Unlike in \citet{kosiorek2019stacked}, there is no direct flow of information from the image to the higher level part of the model; we do not find this to be as important as reported in that work, possibly because we avoid the need for a stop gradient.

\section{Experimental Results}
\label{results}
\subsection{Unsupervised Classification}

The specialisation of parent capsules can be evaluated by classifying inputs based on the presence variables $t^0$. We would expect that, if the generative model has been fit well, the top-level objects roughly correspond to the image classes in the dataset, and so we should be able to classify these with high accuracy based only on this latent representation.

We also want to show that we can perform comparably to previous approaches on this task to validate our claim that our probabilistic formulation corresponds to the assumptions made in previous work on capsules.
However, it is difficult to have an exact performance comparison due to conceptual differences in the interpretation of the model parameters.
\citet{rawlinson2018sparse} perform linear classification on the latent vectors, but these use the latent representation of \citet{sabour2017dynamic}, which does not explicitly attempt to disentangle presence and pose.
\citet{kosiorek2019stacked} perform linear classification and k-means clustering on metrics they refer to as the `prior' and `posterior' capsule presences, which do not correspond directly to posteriors over any variables in our formulation.\footnote{The the closest analogue would be that the `prior' presence is roughly analogous to $t^p$, and the closest analogue to the `posterior' capsule presence would be interpretable as the log of the expected number of child capsules that are coupled to a particular parent. This is discussed in more detail in the supplementary material.}

In order to disambiguate these results, we report linear classification results based on the top-level activation $t^0$ only, as well as linear classification that also include the top-level poses $A^0$, which we find improves performance considerably, bringing our classification results in line with previous work. These results are shown in table \ref{results_rotation_trained_norotation}, along with results on generalising to out of distribution data described in the next section.

\subsection{Generalisation to out of distribution data}

In line with other work on capsules, we wish to show that our model generalises to out of distribution data. In previous work, the standard way to demonstrate this is to evaluate the ability to classify based on the latent representation of a model trained on a standard dataset on new viewpoints. Here, we use the standard benchmark of affNIST \citep{tielemanaffnist}. However, affNIST only includes relatively small changes in viewpoint, so we also train on more challenging out of distribution settings, namely MNIST rotated by up to 180 degrees.

\begin{table}
\begin{center}
\begin{small}
\begin{sc}
\begin{tabular}{ c c c c c| c} \hline 
 \multicolumn{5}{c|}{\textbf{Rotated MNIST}} & {\textbf{affNIST}} \\ 
 $0$ & $45$ & $90$ & $135$ &$180$ & \\ \hline
 $0.96$ & $0.78$ &  $0.56$ & $0.47$ & $0.44$ & $0.88$\\
 $(0.01)$ & $(0.02)$ &$(0.02)$& $(0.01)$ & $(0.01)$ & $(0.03)$\\
  $0.97$ & $0.90$ & $0.77$ & $0.69$ &$0.65$ & $0.92$\\
 $(0.00)$ & $(0.01)$ & $(0.01)$ & $(0.01)$ & $(0.01)$ & $(0.03)$\\
  $0.94$ &  $0.69$  & $0.44$ & $0.35$ & $0.34$& $0.40$\\
 $(0.00)$ &  $(0.01)$ &$(0.01)$  & $(0.01)$ & $(0.00)$ & $(0.01)$ \\
\hline
\end{tabular}
\end{sc}
\end{small}
\caption{Out of distribution performance: linear classification accuracy for model trained on MNIST and evaluated on rotated MNIST and affNIST.  The first row is the classification on $t^0$, the second row is on ${t^0, A^0}$ and the third row on the latent parameters $\mu, \sigma$ of a simple VAE, for a comparison to a non pose-aware model. The values in parentheses are the standard deviations, computed over five runs.}
\label{results_rotation_trained_norotation}
\end{center}
\end{table}

Our affNIST results are better than reported in \citet{rawlinson2018sparse}, and slightly worse than achieved in \citet{kosiorek2019stacked}, but we do not use additional sparsity losses or deformable templates. These results are not state of the art, but they validate our claim that our probabilistic model captures the capsule assumptions, as it shares their robustness to this setting.

\subsection{Training on augmented data}

Increased robustness to changes in viewpoint has always been a motivating example for capsule models. However, this has typically been assessed via experiments like those in the previous section, where a model trained on normal data is tested on augmented data.

However, this is a somewhat artificial scenario.  In general, while we may want to augment the data with additional transformations to encourage robustness, like cropping and augmentation, we cannot necessarily know a priori what augmentations the dataset contains.
If our model structure encourages robustness to \textit{unseen} variations in pose, by explicitly constructing a graphical model that assigns a high likelihood to affine transformations of learned objects, then we might hope that the model will perform even better if we have access to these transformations during training.
In practical scenarios, we may not have any choice about this - for example, if we are training on scenes of real objects, we will presumably have them presented in a variety of random orientations.

\begin{figure}
    \centering
    \includegraphics[width=0.5\textwidth]{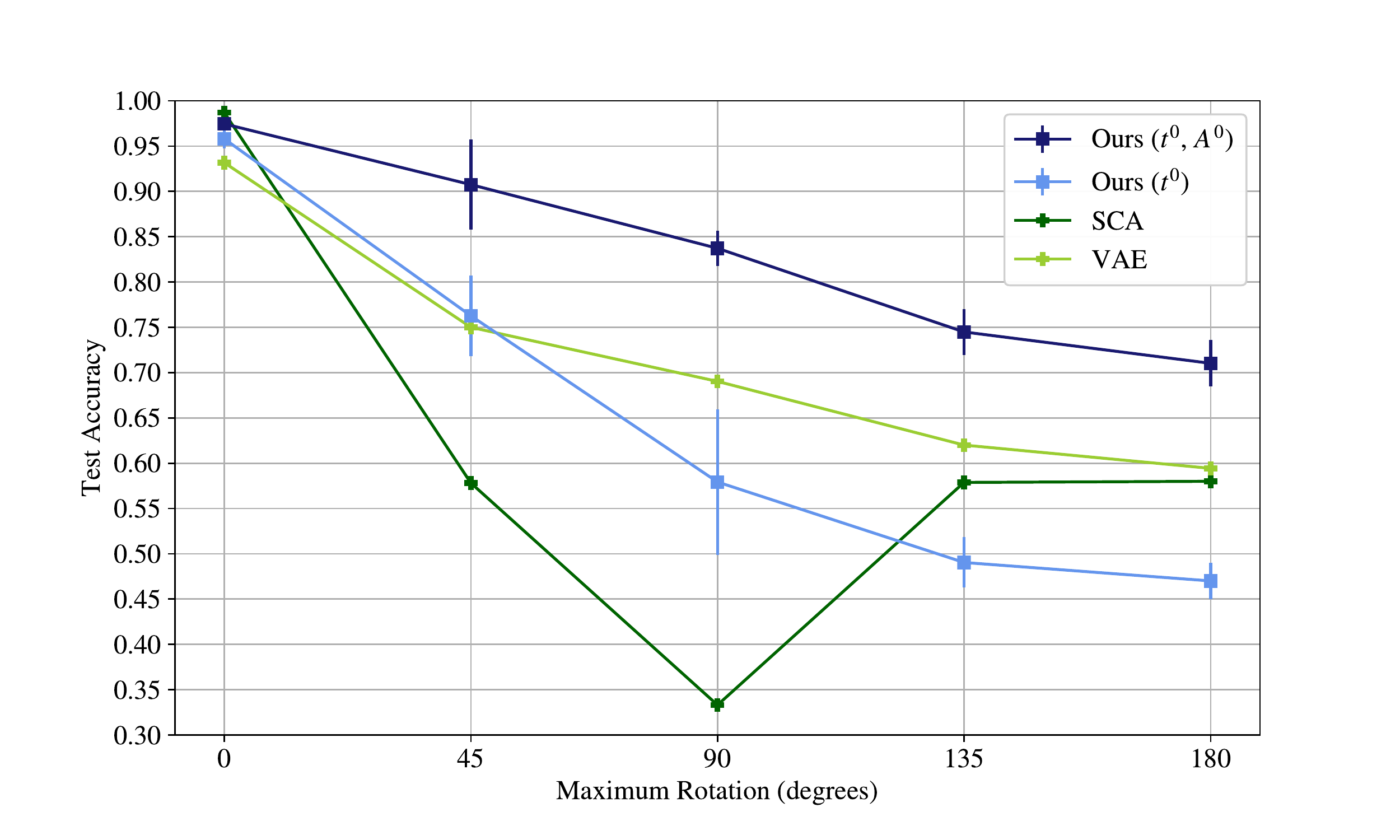}
    \caption{Classification accuracy of latent variables for model trained and tested on Rotated MNIST, compared to a Stacked Capsule Autoencoder (SCA) \citep{kosiorek2019stacked}. We evaluate the degree of specialisation by training a linear classifier on the poses and activations (for the capsule model). In this setting, neither our model using presences only or the model of \citep{kosiorek2019stacked} reliably outperforms a simple VAE, in contrast to the results on \textit{unseen} perturbations in table \ref{results_rotation_trained_norotation}. The results for our models are averaged over five runs. For SCA we use 24 templates as in the original paper; our choice of other hyperparameters is discussed in supplementary material.}
    \label{fig:latent_classification_results}
\end{figure}

Somewhat paradoxically, we find that generative capsule models can perform \textit{worse} in this scenario. We evaluate this by \textit{training} models on rotated MNIST for increasing degrees of rotation.
Despite the latent variables of models trained on normal MNIST being robust to rotations, as shown in the previous section, we find that increased rotations in the training set causes the model to fail to specialise its high-level objects to classes, as shown by the declining performance of classification based on the latent variables. These results are shown in figure \ref{fig:latent_classification_results}.

To verify that this is an issue with this kind of generative model more broadly, and not simply an issue unique to our model formulation, we also perform this experiment on the capsule autoencoder of \citet{kosiorek2019stacked}, using their open source implementation.
The performance of the two models are qualitatively similar, showing that this issue is not specific only to our formulation of capsules. In the next section, we investigate some hypotheses for what causes this failure case.

\begin{figure*}
  \centering
        \includegraphics[ width=0.8\linewidth]{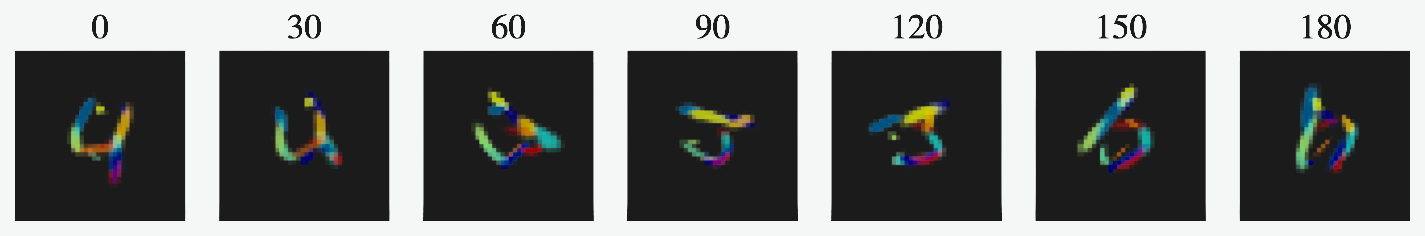}
    \caption{Inference of template arrangement as angular rotation increases. The figures shows the same input image rotated in increments of 30 degrees. Best shown on a computer screen.}
    \label{fig:template_same_image_rotated}
\end{figure*}

\begin{figure}
  \centering
  \includegraphics[width=0.8\columnwidth]{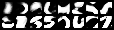}
  \caption{Example learnt templates on MNIST}
  \label{fig:template_example}
\end{figure}

\subsection{Characterising failure modes}

Capsule networks are often evaluated on their generalisation power to unseen perturbations \citep{sabour2017dynamic, hinton2018matrix,kosiorek2019stacked, rawlinson2018sparse}.
However, our results demonstrate a counter-intuitive property---converting these unseen perturbations into \textit{seen} perturbations by adding them to the training set can make it harder, rather than easier, to learn a model like this.

\begin{figure}
\begin{center}
\begin{minipage}{0.41\columnwidth}
        \includegraphics[width=\linewidth]{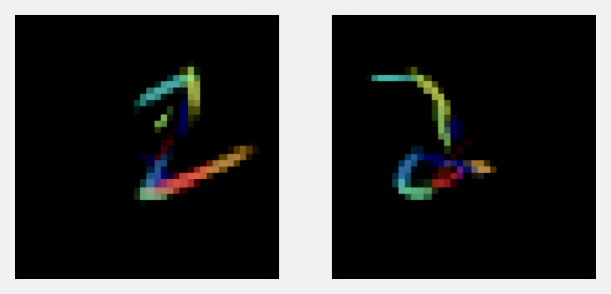}
    \end{minipage}
    \begin{minipage}{0.41\columnwidth}
        \includegraphics[ width=\linewidth]{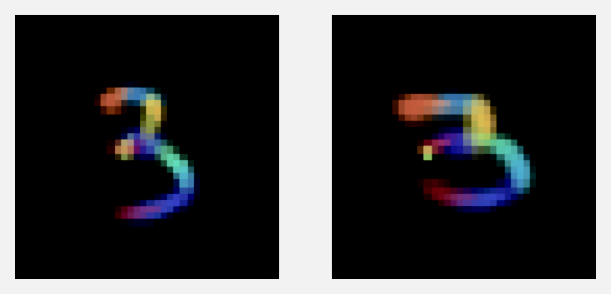}
    \end{minipage}
    \newpage 
    \hspace{0.03cm}
    \begin{minipage}{0.41\columnwidth}
        \includegraphics[width=\linewidth]{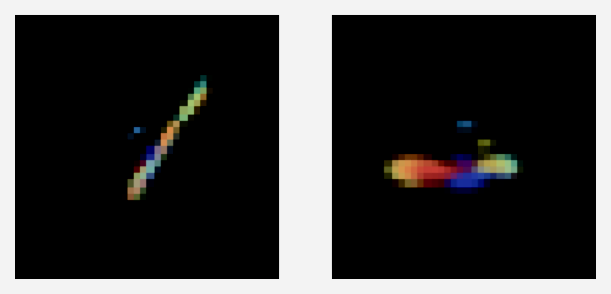}
    \end{minipage}
    \begin{minipage}{0.41\columnwidth}
        \includegraphics[ width=\linewidth]{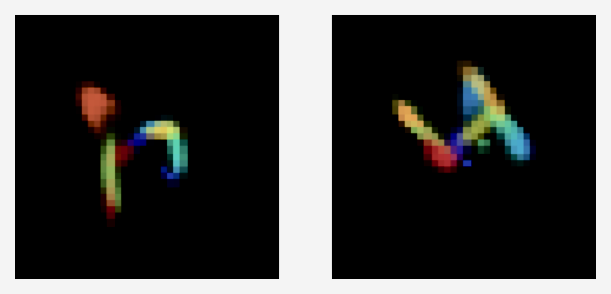}
    \end{minipage}
    \caption{Example of template arrangement for smaller (first row) and larger (second row) angular perturbations. For small angular perturbations, the same template is used for the corresponding stroke in each digit, allowing them to be explained well as a geometrical arrangement of templates with a global transformation applied. For large perturbations, on the other hand, the model uses different templates for the corresponding parts of digits.}
    \label{fig:template_composition}
\end{center} 
\end{figure}

One reason for this is an identifiability problem between parts.
The constraint that there is only a \textit{single} instance of any given object in the model at any one time, which is implicit in the generative model outlined above, often leads to the spurious duplication of parts.
Figure \ref{fig:template_example} shows example templates learned on the MNIST dataset: these are superficially sensible, as they converge to stroke-like shapes.
However, it is clear that many templates are not particularly distinct, and many are nearly affine transformations of each other.

This is a real issue because the templates have an individual identity, as the various parameters $M_{ij}, \rho_{ij}$ are specific to the particular child $i$, so even if the image likelihood is invariant if the poses of two objects are exchanged, the overall likelihood of the model may \textit{not} be.
The assumption in the generative model that a parent can explain a pose as an arrangement of \textit{the same} parts hinges on the same part capsules being consistently used to draw the corresponding part of an object.

However, training time augmentations make this challenging to achieve. This is shown in Figure \ref{fig:template_composition}, which shows reconstructions of digits where the different templates (which have been arbitrarily coloured for illustrative purposes) to show how the image is constructed from the low-level objects. Figure \ref{fig:template_same_image_rotated} shows similar behaviour for the inference of parts when we rotate an input image.
Finding ways to resolve this identifiability issue is resolved is likely an important avenue for future work on this class of model.

\section{Discussion \& Conclusion}

We have presented a probabilistic, generative formulation of capsule networks, encompassing many of the assumptions in previous work.
Our model performs comparably to previous results in the literature on benchmarks they selected, showing that our model is at least a close approximation of these assumptions.
Our probabilistic treatment makes the assumptions of the model explicit and provides a single, unified, consistent objective for learning the model, in contrast to previous work on capsules.

Though we use a fairly straightforward approach to inference in this work, our probabilistic treatment and a unified variational bound would allow experimentation with other inference methods in the future.
For example, other object centred generative
models \cite{greff2019multi} have found it beneficial to use iterative inference at test time, combining traditional and amortised inference approaches.
Our results on test-time free variational inference suggest that this may be a productive direction. Further, recent work on alternative variational objectives to the ELBO \cite{le2018revisiting, masrani2019thermodynamic} for discrete or stochastic support models might be a viable alternative to the concrete relaxation used here. Having a well-defined probabilistic model makes it straightforward conceptually, if not necessarily in practice, to change the inference procedure without modifying the generative model.

Capsule models aim to create explicit representations of objects and their poses. Unlike experimental evaluation in previous work, which has focused on benchmarking performance on tasks like unsupervised learning and generalisation to new viewpoints, we conduct detailed experiments into whether these theoretical properties are, in fact, preserved by this kind of model. Our results show that this kind of generative model may be helpful for enforcing desirable equivariance properties, but that this is far from sufficient.
In addition, our results show that this formulation, while promising, is still in some ways underdetermined.
The issues of identifiability of objects, in particular, suggest that changes to the generative model are likely necessary.

We hope that the probabilistic formulation of this model and our experimental tests of its properties will form a useful basis for future work to build on.

\section*{Acknowlegements}

Lewis Smith is supported by the EPSRC Centre for Doctoral Training in Autonomous Intelligent Machines \& Systems, Grant number EP/L015897/1.
The authors would like to thank A.~Kosiorek for making their code avaliable, and for helpful discussions. We would also like to thank all the members of OATML \& OXCSML for support and feedback.

\FloatBarrier
\bibliography{paper}
\bibliographystyle{icml2020}
\newpage

\onecolumn

\appendix
\icmltitle{Supplementary Material}

\section{Deriving the variational bound}

\newcommand{\Akvar}{A^k_{1:n_k}}
\newcommand{\tkvar}{t^k_{1:n_k}}
\newcommand{\Akcvar}{A^{k+1}_{1:n_{k+1}}}
\newcommand{\tkcvar}{t^{k+1}_{1:n_{k+1}}}
\newcommand{\skcvar}{s^{k+1}_{1:n_{k+1}}}
\newcommand{\Akpvar}{A^{k-1}_{1:n_{k-1}}}
\newcommand{\tkpvar}{t^{k-1}_{1:n_{k-1}}}
\newcommand{\skpvar}{s^{k-1}_{1:n_{k-1}}}
\newcommand{\skcivar}{s^{k+1}_{i}}
\newcommand{\Akcivar}{A^{k+1}_{i}}
\newcommand{\tkcivar}{t^{k+1}_{i}}
\newcommand{\tallvar}{t^{0:N}_{1:n_k}}
\newcommand{\Aallvar}{A^{0:N}_{1:n_k}}

In the main body, the full derivation of the ELBO was omitted,  but the independence assumptions and form of the bound are discussed. Here we provide a full derivation.

We wish to approximate the posterior $p(\tallvar, \Aallvar \mid X)$. Here the upper index is into layers of random variables, and the lower index is over capsules in that layer. 
Although we only use two `hidden layers' in this work, we derive a general bound, as it is little extra effort. 

Note, as discussed in the text, we do not apply a variational distribution over $s$, since the conditional independence assumptions are such that it can be summed out of the variational bound.
In more detail, conditional on $\Akvar, \tkvar, \Akcvar, \tkcvar$, the activations and poses of the parents, the $\skcivar$ become conditionally independent of each other.

This will appear in the bound implicitly, as it lets us calculate 

\begin{align}
    p(\Akcivar, \tkcivar \mid \Akvar, \tkvar) &= \sum_{j \in n_{k}} p(\Akcivar, \tkcivar \mid A^{k}_j, t_j^{k}, \skcivar=j) p(\skcivar=j \mid A^{k}_j, t_j^{k}) 
\end{align}

Implicitly, this corresponds to an `E' step in EM; since we will take a gradient step to maximise variational distribution over the $t$s and $A$s holding the implicit `soft assignment' over the discrete variables constant.

The variational bound takes the form 
 
\begin{align*}
	\mathcal{L}[q] = &\expect{q(\tallvar, \Aallvar)}{\log p(\tallvar, \Aallvar, X) - \log q(\tallvar, \Aallvar)}
\end{align*}

In order to make it clear how this can be computed, we note it can be split up layerwise, using the factorisation of the graphical model that each set of random variables depends only on it's immediate parents

\begin{align*}
    \log p(\tallvar, \Aallvar, X) &= \log p(X \mid t^N_{1:{n_N}}, A^N_{1:{n_N}}) \\
    +& \log p(t^{N}_{1:n_{N}}, A^N_{1:n_{N}} \mid t^{N-1}_{1:n_{N-1}}, A^{N-1}_{1:n_{N-1}}) \\
    \vdots\\
    +&\log p(\tkcvar, \Akcvar \mid \tkvar, \Akvar) \\
    \vdots \\
    +& \log p(t^{1}_{1:n_{1}}, A^1_{1:n_{1}} \mid t^{0}_{1:n_{0}}, A^{0}_{1:n_{0}}) \\
    +& \log p(t^{0}_{1:n_{0}}, A^{0}_{1:n_{0}})
\end{align*}

We assume a similar layerwise structure for the variational posterior, so that $q$ factorises as

\begin{align*}
    \log q(\tallvar, \Aallvar) &= \log q(t^N_{1:{n_N}}, A^N_{1:{n_N}} \mid X)\\
    +& \log q(t^{N-1}_{1:n_{N-1}}, A^{N-1}_{1:n_{N-1}} \mid t^{N}_{1:n_{N}}, A^N_{1:n_{N}} ) \\
    \vdots\\
    +&\log q(\tkvar, \Akvar \mid \tkcvar, \Akcvar) \\
    \vdots\\
    +& \log q(t^{0}_{1:n_{0}}, A^{0}_{1:n_{0}} \mid t^{1}_{1:n_{1}}, A^1_{1:n_{1}}) \\
\end{align*}

that is, we assume that the posterior over the random variables at layer $k$ only depends on the variables in the layer immediately below. We also make a mean field assumption within a layer, that is, we assume $q(\tkvar, \Akvar \mid ...) = \prod q(t^k_i \mid ...) q(A^k_i\mid ...)$, but this is not critical for the following derivation and clutters notation, so we omit it from consideration for now.

We can therefore group terms in the variational bound by layer as follows

\begin{align*}
    \log p(\tallvar, \Aallvar, X) -\log q(\tallvar, \Aallvar) &= \log p(X \mid t^N_{1:{n_N}}, A^N_{1:{n_N}})\\
    +& \log p(t^{N}_{1:n_{N}}, A^N_{1:n_{N}} \mid t^{N-1}_{1:n_{N-1}}, A^{N-1}_{1:n_{N-1}}) - \log q(t^N_{1:{n_N}}, A^N_{1:{n_N}} \mid X)\\
    \vdots\\
    +&\log p(\tkvar, \Akvar \mid \tkpvar, \Akpvar) - \log q(\tkvar, \Akvar \mid \tkcvar, \Akcvar) \\
    \vdots \\
    +& \log p(t^{0}_{1:n_{0}}, A^{0}_{1:n_{0}}) - \log q(t^{0}_{1:n_{0}}, A^{0}_{1:n_{0}} \mid t^{1}_{1:n_{1}}, A^1_{1:n_{1}}) \\
\end{align*}

Notice that, apart from the likelihood term $p(X \mid t^N_{1:{n_N}}, A^N_{1:{n_N}})$, these terms all take the from of relative entropies $\kl{p}{q} = \expect{q}{\log p(x) - \log q(x)}$.

A MC estimator of the evidence lower bound could therefore be computed in a layerwise fashion, first sampling from $q(t^N_{1:{n_N}}, A^N_{1:{n_N}} \mid X)$, computing the likelihood term and the KL term for that layer, then computing $q(t^{N-1}_{1:{n_N}}, A^{N-1}_{1:{n_N}} \mid X)$, and proceeding layerwise upwards, as only the variables in the layer immediately above or below are ever needed to compute all of these terms.

In practice, as mentioned in the text, we use a bound estimator of this form automatically derived by Pyro's tracing mechanics rather than keeping track of all this bookkeeping manually in our code.
This means in practice that we use stochastic estimators of the KL terms even when in principle deterministic expressions could be used, as Pyro does not currently support analytic KL divergences.

\section{Model implementation details}

Here we detail some more mundane details of the model. These could be found by reading the code, to be released, but we record them for greater reproducibility.

The model has a number of hyperparameters, other than the learning rates and neural network hyperparameters and the number of capsules to use in each layer.

The use of Concrete distributions over the $t$ variables introduces a temperature hyperparamter.
In addition, the standard deviation parameters $c_{ij}$ and the standard deviation of the pixel noise $\sigma$ have to be constrained - allowing these to be learned freely allows increasing terms in the variational bound without limit.
A more general solution would be to place hyperpriors over these parameters, but for simplicity we reparameterise these so that there is a minimum value each can take, which is a hyperparameter. The minimum $c$ and $\sigma$ and the temperature are fairly important, and modifying them has a fairly significant effect on performance. It is possible that the training of the model could be improved by, for example, using a temperature schedule, but we did not experiment with this.

The binding probability $\rho$ of the dummy parent and the width $\lambda_{off}$ of inactive poses are hyperparameters.
In our experiments, the model seemed relatively insensitive to these.
We set the pixel standard deviation to $0.2$. We learn the covariance noise, but set it to have a minimum value of $0.1$. We use a temperature of $1.0$ for all relaxed variables.

For the convnet that maps from the image to the lowest level latent variables, we use 2 convolutional layers with kernel size $3$ and a stride of $2$, followed by $2$ convolutional layers with kernel size $3$ and stride $1$ and attentive pooling. We use ELU nonlinearities.

For the set transformer, we use $4$ ISAB layers with $128$ hidden units, $32$ inducing points and $4$ attention heads, with layer norm, followed by a PMA block and a linear layer to a $32$ dimensional encoding.
Independent one layer MLPs with $128$ hidden units are used to map from this encoding to the parameters of all variational distributions.

\section{Details of SCAE Hyperparameters}

\citet{kosiorek2019stacked} report using $24$ object and part capsules for MNIST, and setting all the auxiliary losses to 1 except the posterior losses, which are set to 10.
However, in the public code release (\url{https://github.com/google-research/google-research/tree/master/stacked_capsule_autoencoders}), the default script for MNIST uses different hyperparameter settings for these losses than reported in the paper, and 40 object and part capsules. Namely,
the \texttt{run\_mnist} script uses  a posterior between example sparsity of 0.2, 
  a posterior within example sparsity weight of 0.7, 
  a prior between example sparsity weight of 0.35, 
  a prior within example constant 4.3 and a
  a prior within example sparsity weight of 2.

For the SCAE results reported in the paper, we use 24 capsules as in \citet{kosiorek2019stacked}, but we use the loss settings from the code repository, since we tried both and found these to work slightly better.
We do not use more capsules because we find this could lead to confusion with the main point of this experiment, which is to test the degree to which the model is capable of separating out pose from visual appearance - having more capsules allows a great deal of redundancy, for example, learning several different objects for a 2 rotated by various degrees, rather than recognising that an image can be explained as a single object in different orientations.
As rotated MNIST is identical to normal MNIST apart from the variation in object orientation, it should not in principle require more capsules to explain this data with the generative model.

\section{SCAE: Prior and Posterior Presences}

\citet{kosiorek2019stacked} report two ways of quantifying the presence of object capsules - the `prior' presence $a_k \max_m a_{k, m}$ (in the notation of that paper) and the `posterior' presence $\sum_m a_k a_{k, m} \mathcal{N}(x_m, k, m)$. We report results above with the posterior presence, as in the paper, which performs slightly better.
In \citet{kosiorek2019stacked}, these are not explicitly modelled as random variables. The model described in our paper does not correspond exactly to that in \citet{kosiorek2019stacked}, as discussed, but if analogies are drawn, then the $a_k$ correspond to $t$, and the parameters $a_{k, m}$ correspond to the parent child affinities $\rho$ (we treat these as constant, whereas these are modifiable per input in that work).

The `prior' probability is roughly equivalent to our use of the logit of the posterior over $t_0$ for classification. The posterior presence has no immediately obvious correspondence to the posterior over any random variables in our model, but it can be interpreted as the expected number of child capsules that couple to a parent. This can be seen as follows;

The posterior over $s$ conditional on the parent and child poses and presences can be straightforwardly calculated using Bayes rule:

\begin{align*}
p(s_j = i \mid t_j^1, A_j^1, t_{1:n}^0, A_{1:n}^0) \propto p(t^1_j, A^1_j \mid s_j=i, t_{1:n}^0, A_{1:n}^0) p(s=i \mid t_{1:n}^0, A_{1:n}^0)
\end{align*}

That is to say, the posterior probability of child $j$ attaching to parent $i$ is given by the above quantity.
If we sum this over the children, then we get a very similar expression to the `posterior presence' score, ignoring the terms that would correspond to the child presence, which is not assigned a likelihood in the model of \citet{kosiorek2019stacked}.
This sum over the children is the expected number of children attached to parent $i$ under the posterior distribution.

\end{document}